\documentclass[letterpaper]{SprMixLaTeX}

\usepackage{cite}
\usepackage{graphicx}
\usepackage[figuresright]{rotating}

\begin{document}

\title{Guiding Designs of Self-Organizing Swarms: Interactive and Automated Approaches}
\titlerunning{Guiding Designs of Self-Organizing Swarms}

\author{Hiroki Sayama}

\institute{Collective Dynamics of Complex Systems Research Group\\
Binghamton University, State University of New York\\
Binghamton, NY 13902-6000, USA\\
sayama@binghamton.edu}

\maketitle

\begin{abstract}
Self-organization of heterogeneous particle swarms is rich in its
dynamics but hard to design in a traditional top-down manner,
especially when many types of kinetically distinct particles are
involved. In this chapter, we discuss how we have been addressing this
problem by (1) utilizing and enhancing interactive evolutionary design
methods and (2) realizing spontaneous evolution of self-organizing
swarms within an artificial ecosystem. \footnote{This chapter is based
  on our previous publications
  \cite{sayama07,sayama09,sayama09ieee,sayama10,bush11,sayama11a,sayama11b,sayama12}.}
\end{abstract}

\vspace{0.7 cm}

\section{Introduction}

Engineering design has traditionally been a top-down process in which
a designer shapes, arranges and combines various components in a
specific, precise, hierarchical manner, to create an artifact that
will behave deterministically in an intended way
\cite{minai2006complex,pahl2007engineering}. However, this process
does not apply to complex systems that show self-organization,
adaptation and emergence.  Complex systems consist of a massive amount
of simpler components that are coupled locally and loosely, whose
behaviors at macroscopic scales emerge partially stochastically in a
bottom-up way. Such emergent properties of complex systems are often
very robust and dynamically adaptive to the surrounding environment,
indicating that complex systems bear great potential for engineering
applications \cite{ottino2004engineering}.
 
\begin{figure}[tbp]
\centering
\includegraphics[width=\textwidth]{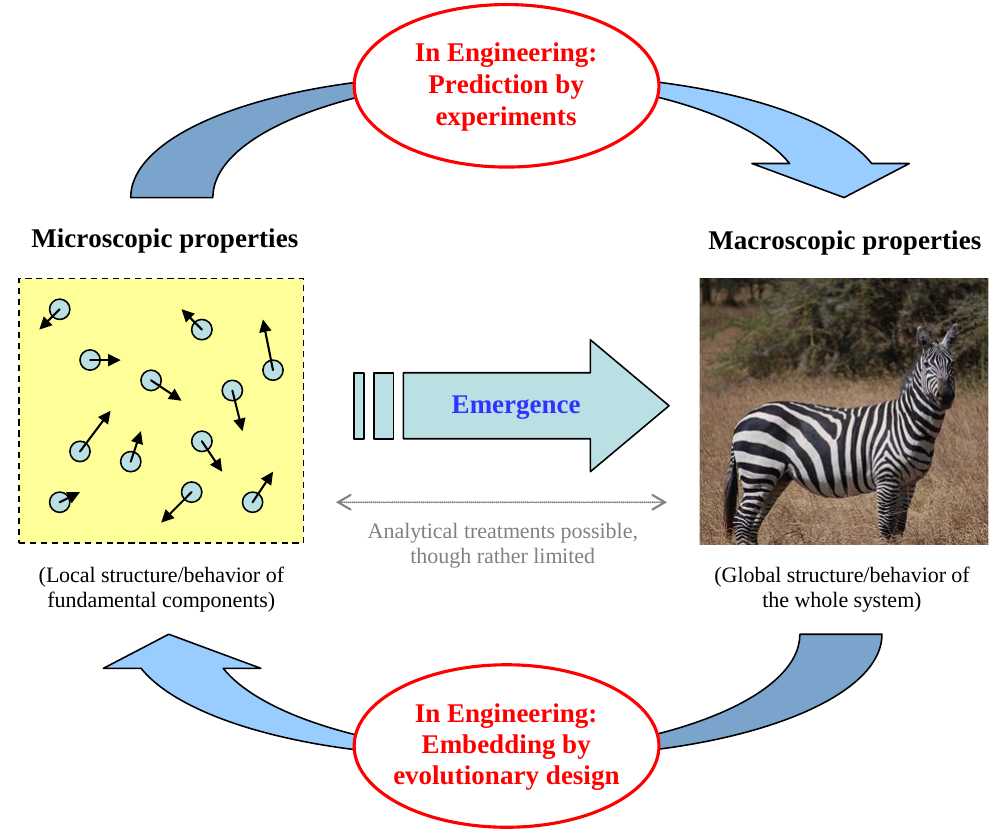}
\caption{Relationships of macroscopic and microscopic properties in
  complex systems and how complex systems engineering has been
  handling the gap between them.}
\label{fig1}
\end{figure}

In an attempt to design engineered complex systems, one of the most
challenging problems has been how to bridge the gap between macro and
micro scales. Some mathematical techniques make it possible to
analytically show such macro-micro relationships in complex systems
(e.g., those developed in statistical mechanics and condensed matter
physics \cite{bar2003dynamics,boccara2010modeling}).  However, those
techniques are only applicable to ``simple'' complex systems, in
which: system components are reasonably uniform and homogeneous, their
interactions can be approximated without losing important dynamical
properties, and/or the resulting emergent patterns are relatively
regular so that they can be characterized by a small number of
macroscopic order parameters
\cite{bar2003dynamics,MEbookChap1}. Unfortunately, such cases are
exceptions in a vast, diverse, and rather messy compendium of complex
systems dynamics \cite{camazine2003self,sole2008signs}. To date, the
only generalizable methodology available for predicting macroscopic
properties of a complex system from microscopic rules governing its
fundamental components is to conduct experiments---either
computational or physical---to let the system show its emergent
properties by itself (Fig. \ref{fig1}, top).

More importantly, the other way of connecting the two
scales---embedding macroscopic requirements the designer wants into
microscopic rules that will collectively achieve those
requirements---is by far more difficult. This is because the mapping
between micro and macro scales is highly nonlinear, and also the space
of possible microscopic rules is huge and thus hard to explore. So
far, the only generalizable methodology available for macro-to-micro
embedding in this context is to acquire microscopic rules by
evolutionary means \cite{bentley1999evolutionary} (Fig. \ref{fig1},
bottom). Instead of trying to derive local rules analytically from
global requirements, evolutionary methods let better rules
spontaneously arise and adapt to meet the requirements, even though
they do not produce any understanding of the macro-micro
relationships. The effectiveness of such ``blind'' evolutionary search
\cite{dawkins1996blind} for complex systems design is empirically
supported by the fact that it has been the primary mechanism that has
produced astonishingly complex, sophisticated, highly emergent
machinery in the history of real biological systems.

The combination of these two methodologies---experiment and
evolution---that connect macro and micro scales in two opposite
directions (the whole cycle in Fig. \ref{fig1}) is now a widely adopted
approach for guiding systematic design of self-organizing complex
systems \cite{minai2006complex,anderson2006creation}. Typical design
steps are to (a) create local rules randomly or using some heuristics,
(b) conduct experiments using those local rules, (c) observe what kind
of macroscopic patterns emerge out of them, (d) select and modify
successful rules according to the observations, and (e) repeat these
steps iteratively to achieve evolutionary improvement of the
microscopic rules until the whole system meets the macroscopic
requirements.

Such experiment-and-evolution-based design of complex systems is not
free from limitations, however. In typical evolutionary design
methods, the designer needs to explicitly define a performance metric,
or ``fitness'', of design candidates, i.e., how good a particular
design is. Such performance metrics are usually based on relatively
simple observables easily extractable from experimental results (e.g.,
the distance a robot traveled, etc.). However, simple quantitative
performance metrics may not be suitable or useful in evolutionary
design of more complex structures or behaviors, such as those seen in
real-world biological systems, where the key properties a system
should acquire could be very diverse and complex, more qualitative
than quantitative, and/or even unknown to the designer herself
beforehand.

In this chapter, we present our efforts to address this problem, by
(1) utilizing and enhancing interactive evolutionary design methods
and (2) realizing spontaneous evolution of self-organizing swarms
within an artificial ecosystem.

\section{Model: Swarm Chemistry}

\begin{figure}[tbp]
\centering
\includegraphics[width=0.7\columnwidth]{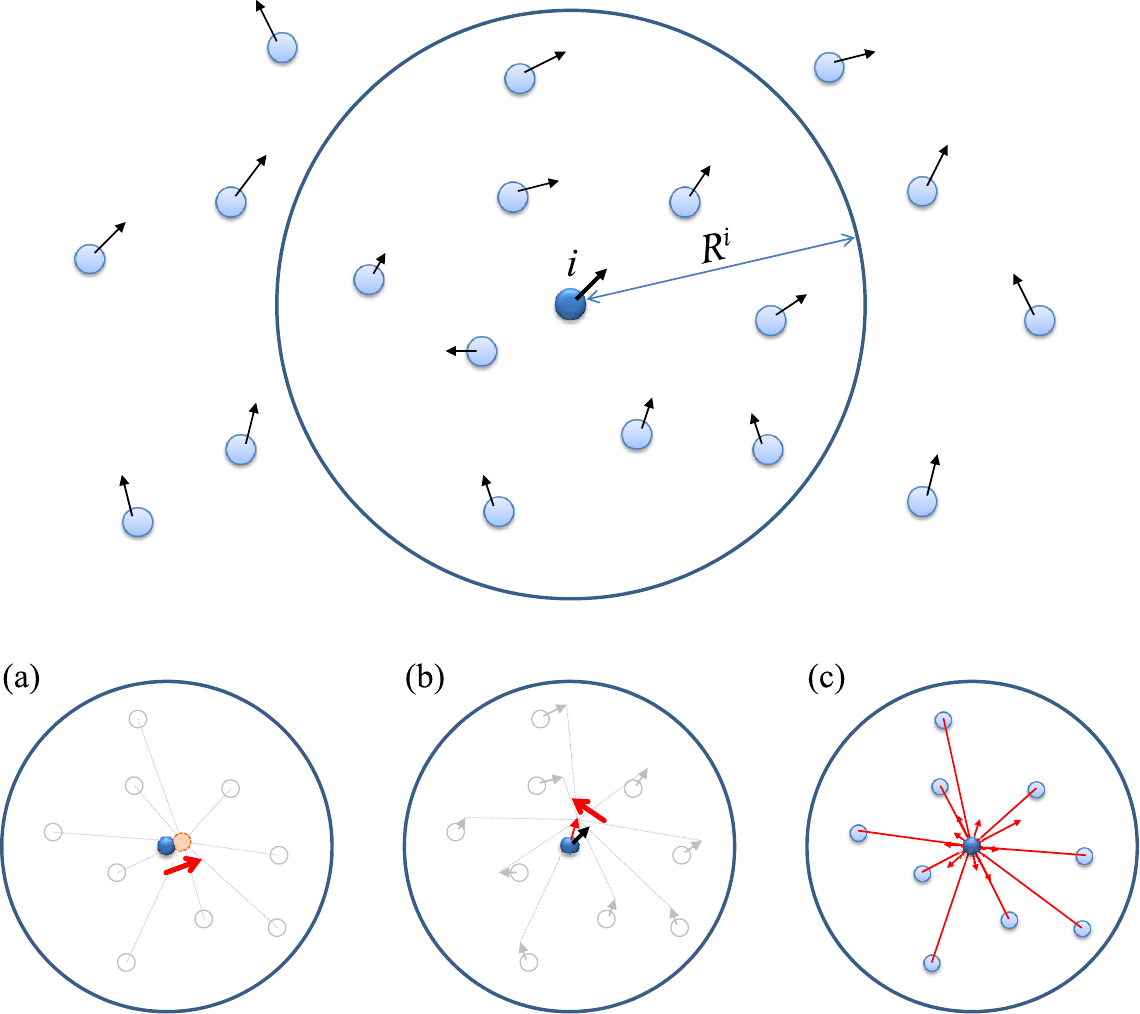}
\caption{Kinetic interactions between particles (from
  \cite{sayama10}). Top: Particle $i$ senses only positions and
  velocities of nearby particles within distance $R^i$. Bottom: (a)
  Cohesion. Particle $i$ accelerates toward the center of mass of
  nearby particles. (b) Alignment. Particle $i$ steers to align its
  orientation to the average orientation of nearby particles. (c)
  Separation. Particle $i$ receives repulsion forces from each of the
  nearby particles whose strength is inversely related to distance.}
\label{interaction}
\end{figure}

We use Swarm Chemistry \cite{sayama07,sayama09} as an example of
self-organizing complex systems with which we demonstrate our design
approaches. Swarm Chemistry is an artificial chemistry
\cite{dittrich01} model for designing spatio-temporal patterns of
kinetically interacting heterogeneous particle swarms using
evolutionary methods. A swarm population in Swarm Chemistry consists
of a number of simple particles that are assumed to be able to move to
any direction at any time in a two- or three-dimensional continuous
space, perceive positions and velocities of other particles within its
local perception range, and change its velocity in discrete time steps
according to the following kinetic rules (adopted and modified from
the rules in Reynolds' Boids \cite{reynolds87}; see
Fig.~\ref{interaction}):
\begin{itemize}
\item If there are no other particles within its local perception
  range, steer randomly ({\em Straying}).
\item Otherwise:
\begin{itemize}
\item Steer to move toward the average position of nearby particles
  ({\em Cohesion}, Fig.~\ref{interaction}(a)).
\item Steer toward the average velocity of nearby particles ({\em
  Alignment}, Fig.~\ref{interaction}(b)).
\item Steer to avoid collision with nearby particles ({\em
  Separation}, Fig.~\ref{interaction}(c)).
\item Steer randomly with a given probability ({\em Randomness}).
\end{itemize}
\item Approximate its speed to its own normal speed ({\em
  Self-propulsion}).
\end{itemize}
These rules are implemented in a simulation algorithm that uses
kinetic parameters listed and explained in Table \ref{parameters} (see
\cite{sayama09,sayama10} for details of the algorithm). The kinetic
interactions in our model uses only one omni-directional perception
range ($R^i$), which is much simpler than other typical swarm models
that use multiple and/or directional perception ranges
\cite{reynolds87,couzin02,kunz03,hemelrijk05,cheng05,newman08}. Moreover,
the information being shared by nearby particles is nothing more than
kinetic one (i.e., relative position and velocity), which is
externally observable and therefore can be shared without any
specialized communication channels\footnote{An exception is local
  information transmission during particle recruitment processes,
  which will be discussed later.}. These features make this system
uniquely simple compared to other self-organizing swarm models.

\begin{table}[tbp]
\caption{Kinetic parameters involved in the simulation of particle
  behavior (from \cite{sayama10}). Unique values are assigned to these
  parameters for each particle $i$ as its own kinetic properties.}
\label{parameters}
\centering
\begin{tabular}{crrlc}
\hline
Name & Min & Max & \multicolumn{1}{c}{Meaning} & Unit\\
\hline
$R^i$ & 0 & 300 & Radius of local perception range & pixel\\
$V^i_n$ & 0 & 20 & Normal speed & pixel step$^{-1}$\\
$V^i_m$ & 0 & 40 & Maximum speed & pixel step$^{-1}$\\
$c^i_1$ & 0 & 1 & Strength of cohesive force & step$^{-2}$\\
$c^i_2$ & 0 & 1 & Strength of aligning force & step$^{-1}$\\
$c^i_3$ & 0 & 100 & Strength of separating force & pixel$^2$ step$^{-2}$\\
$c^i_4$ & 0 & 0.5 & Probability of random steering & ---\\
$c^i_5$ & 0 & 1 & Tendency of self-propulsion & ---\\
\hline
\end{tabular}
\end{table}

Each particle is assigned with its own kinetic parameter settings that
specify preferred speed, local perception range, and strength of each
kinetic rule. Particles that share the same set of kinetic parameter
settings are considered of the same type. Particles do not have a
capability to distinguish one type from another; all particles look
exactly the same to themselves.

For a given swarm, specifications for its macroscopic properties are
indirectly and implicitly woven into a list of different kinetic
parameter settings for each swarm component, called a {\em recipe}
(Fig.~\ref{recipe}) \cite{sayama09}. It is quite difficult to manually
design a specific recipe that produces a desired structure and/or
behavior using conventional top-down design methods, because the
self-organization of a swarm is driven by complex interactions among a
number of kinetic parameters that are intertwined with each other in
highly non-trivial, implicit ways.

\begin{figure}[tbp]
\centering
\fbox{\parbox[b]{0.65\columnwidth}{
97 * (226.76, 3.11, 9.61, 0.15, 0.88, 43.35, 0.44, 1.0)\\
38 * (57.47, 9.99, 35.18, 0.15, 0.37, 30.96, 0.05, 0.31)\\
56 * (15.25, 13.58, 3.82, 0.3, 0.8, 39.51, 0.43, 0.65)\\
31 * (113.21, 18.25, 38.21, 0.62, 0.46, 15.78, 0.49, 0.61)
}}
\caption{Example of a recipe, formatted as a list of kinetic parameter
  sets of different types within a swarm (from \cite{sayama10}). Each
  row represents one type, which has a number of particles of that
  type at the beginning, followed by its parameter settings in the
  format of $(R^i, V^i_n, V^i_m, c^i_1, c^i_2, c^i_3, c^i_4, c^i_5)$.}
\label{recipe}
\end{figure}

In the following sections, we address this difficult design problem
using evolutionary methods. Unlike in other typical evolutionary
search or optimization tasks, however, in our swarm design problem,
there is no explicit function or algorithm readily available for
assessing the quality (or fitness) of each individual design. To meet
with this unique challenge, we used two complementary approaches: The
interactive approach, where human users are actively involved in the
evolutionary design process, and the automated approach, where
spontaneous evolutionary dynamics of artificial ecosystems are
utilized as the engine to produce creative self-organizing patterns.

\section{Interactive Approach}

The first approach is based on interactive evolutionary computation
(IEC) \cite{banzhaf2000interactive,takagi2001interactive}, a
derivative class of evolutionary computation which incorporates
interaction with human users. Most IEC applications fall into a
category known as ``narrowly defined IEC'' (NIEC)
\cite{takagi2001interactive}, which simply outsources the task of
fitness evaluation to human users. For example, a user may be
presented with a visual representation of the current generation of
solutions and then prompted to provide fitness information about some
or all of the solutions. The computer in turn uses this fitness
information to produce the next generation of solutions through the
application of a predefined sequence evolutionary operators. 

\begin{figure}[tbp]
\centering
\begin{tabular}{ccc}
``swinger'' & ``rotary'' & ``walker-follower'' \\
\fbox{\includegraphics[width=23mm]{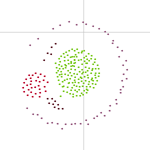}} &
\fbox{\includegraphics[width=23mm]{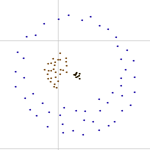}} &
\fbox{\includegraphics[width=23mm]{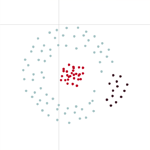}} \\
\end{tabular}
\caption{Examples of swarms designed using IEC methods. Their recipes
  are available on the Swarm Chemistry website
  (http://bingweb.binghamton.edu/\~{}sayama/SwarmChemistry/).}
\label{patterns}
\end{figure}

Our initial work, Swarm Chemistry 1.1 \cite{sayama07,sayama09}, also
used a variation of NIEC, called Simulated Breeding \cite{unemi}. This
NIEC-based application used discrete, non-overlapping generation
changes. The user selects one or two favorable swarms out of a fixed
number of swarms displayed, and the next generation is generated out
of them, discarding all other unused swarms. Selecting one swarm
creates the next generation using perturbation and mutation. Selecting
two swarms creates the next generation by mixing them together
(similar to crossover, but this mixing is not genetic but
physical). Figure \ref{patterns} shows some examples of
self-organizing swarms designed using Swarm Chemistry 1.1.

\begin{figure}[tbp]
\centering
\begin{tabular}{ccc}
$\vcenter{\hbox{\includegraphics[width=0.38\textwidth]{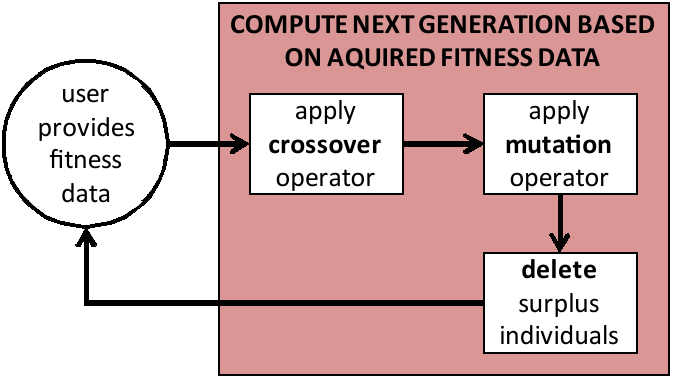}}}$& ~ &
$\vcenter{\hbox{\includegraphics[width=0.58\textwidth]{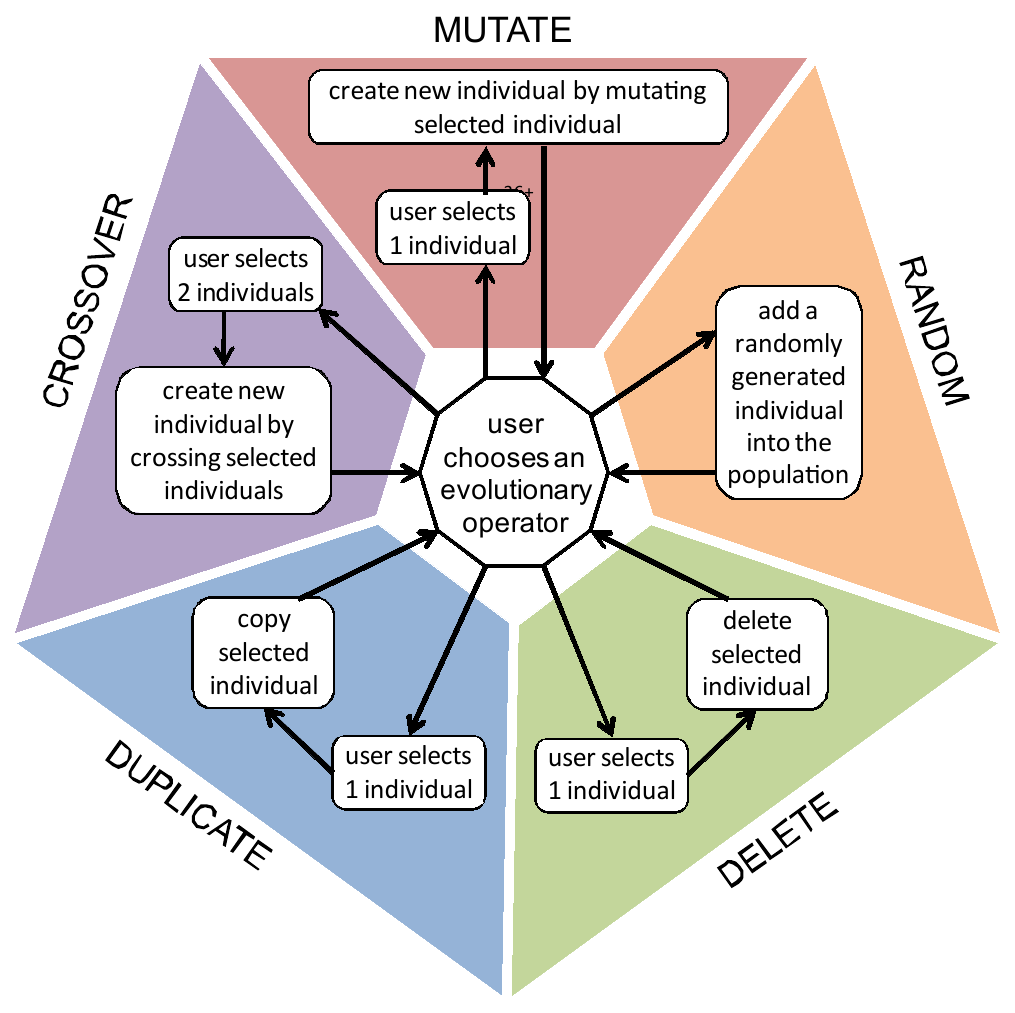}}}$ \\
{\bf NIEC} & ~ & {\bf HIEC} \\
\end{tabular}
\caption{Comparison of control flows between two interactive
  evolutionary computation (IEC) frameworks (from
  \cite{bush11}). Left: Narrowly defined IEC (NIEC). Right:
  Hyper-interactive IEC (HIEC).}
\label{niec-hiec}
\end{figure}

As a design tool, NIEC has some disadvantages. One set of disadvantage
stems from the confinement of the user to the role of selection
operator (Fig. \ref{niec-hiec}, left). Creative users who are
accustomed to a more highly involved design process may find the
experience to be tedious, artificial, and frustrating. Earlier
literature suggests that it is important to instill in the user a
strong sense of control over the entire evolutionary process
\cite{bentley} and that the users should be the initiators of actions
rather than simply responding to prompts from the system
\cite{shneiderman}.

These lines of research suggest that enhancing the level of
interaction and control of IEC may help the user better guide the
design process of self-organizing swarms. Therefore, we developed
the concept of hyperinteractive evolutionary computation (HIEC)
\cite{bush11}, a novel form of IEC in which a human user actively
chooses when and how to apply each of the available evolutionary
operators, playing the central role in the control flow of
evolutionary search processes (Fig. \ref{niec-hiec}, right). In HIEC,
the user directs the overall search process and initiates actions by
choosing when and how each evolutionary operator is applied. The user
may add a new solution to the population through the crossover,
mutate, duplicate, or random operators. The user can also remove
solutions with the delete operator. This naturally results in dynamic
variability of population size and continuous generation change (like
steady-state strategies for genetic algorithms).

We developed Swarm Chemistry 1.2 \cite{sayama09ieee,bush11}, a
redesigned HIEC-based application for designing swarms. This version uses
continuous generation changes, i.e., each evolutionary operator is
applied only to part of the population of swarms on a screen without
causing discrete generation changes. A mutated copy of an existing
swarm can be generated by either selecting the ``Mutate'' option or
double-clicking on a particular swarm. Mixing two existing swarms can
be done by single-clicking on two swarms, one after the other. The
``Replicate'' option creates an exact copy of the selected swarm next
to it. One can also remove a swarm from the population by selecting
the ``Kill'' option or simply closing the frame. More details of HIEC
and Swarm Chemistry 1.2 can be found elsewhere
\cite{sayama09ieee,bush11}.

We conducted the following two human-subject experiments to see if
HIEC would produce a more controllable and positive user experience,
and thereby better swarm design outcomes, than those with NIEC.

\subsection{User experience}

In the first experiment, individual subjects used the NIEC and HIEC
applications mentioned above to evolve aesthetically pleasing
self-organizing swarms. We quantified user experience outcomes using
questionnaire, in order to quantify potential differences in user
experience between the two applications.

Twenty-one subjects were recruited from students and faculty/staff
members at Binghamton University. Each subject was recruited and
participated individually. The subject was told to spend five minutes
using each of two applications to design an ``interesting and
lifelike'' swarm. Each of these two applications ran on their own
dedicated computer station. After completing two sessions, each of
which used either NIEC or HIEC application, the subject filled out a
survey, rating each of the two platforms on the following factors:
easiness of operation, controllability, intuitiveness, fun factor,
fatigue level, final design quality, and overall satisfaction. Each
factor was rated on a 5-point scale.

\begin{figure}[tbp]
\centering
\includegraphics[width=\textwidth]{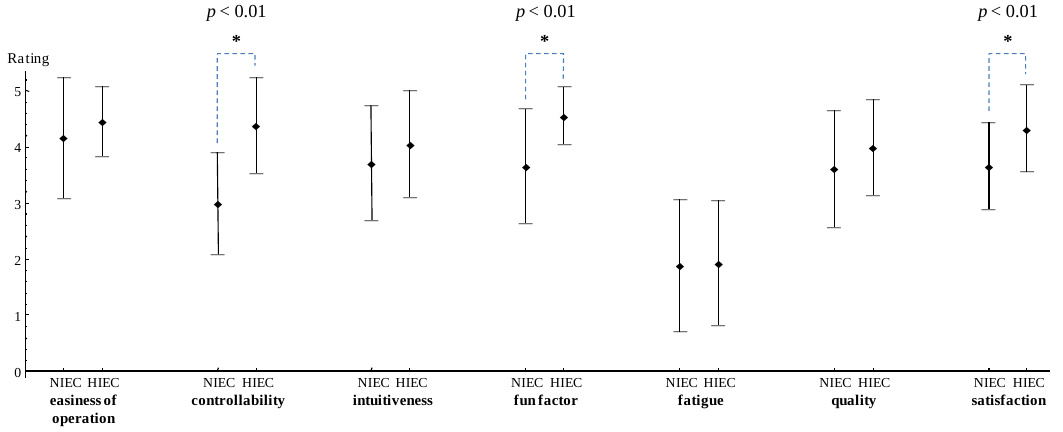}
\caption{Comparison of rating distribution between the NIEC and HIEC
  applications across seven factors. Mean ratings are shown by
  diamonds, with error bars around them showing standard
  deviations. Significant differences are indicated with an asterisk
  and corresponding $t$-test $p$-values.}
\label{userexperience}
\end{figure}

The results are shown in Fig. \ref{userexperience}. Of the 7 factors
measured, 3 showed statistically significant difference between two
platforms: controllability, fun factor, and overall satisfaction. The
higher controllability ratings for HIEC suggest that our original
intention to re-design an IEC framework to grant greater control to
the user was successful. Our results also suggest that this increased
control may be associated with a more positive user experience, as is
indicated by the higher overall satisfaction and fun ratings for
HIEC. In the meantime, there was no significant difference detected in
terms of perceived final design quality. This issue is investigated in
more detail in the following second experiment.

\subsection{Design quality}

The goal of the second experiment was to quantify the difference
between HIEC and NIEC in terms of final design quality. In addition,
the effects of mixing and mutation operators on the final design
quality were also studied. The key feature of this experiment was that
design quality was rated not individually by the subjects who designed
them, but by an entire group of individual subjects. The increased
amount of rating information yielded by this procedure allowed us to
more effectively detect differences in quality between designs created
using NIEC and designs created using HIEC.

Twenty-one students were recruited for this experiment. Those subjects
did not have any overlap with the subjects of experiment 1. The
subjects were randomly divided into groups of three and instructed to
work together as a team to design an ``interesting'' swarm design in
ten minutes using either the NIEC or HIEC application, the latter of
which was further conditioned to have the mixing operator, the
mutation operator, or both, or none. The sessions were repeated so
that five to seven swarm designs were created under each
condition. Once the sessions were over, all the designs created by the
subjects were displayed on a large screen in the experiment room, and
each subject was told to evaluate how ``cool'' each design was on a
0-to-10 numerical scale. Details of the experimental procedure and
data analysis can be found elsewhere \cite{sayama09ieee,bush11}.

\begin{figure}[tbp]
\centering
\includegraphics[width=.65\columnwidth]{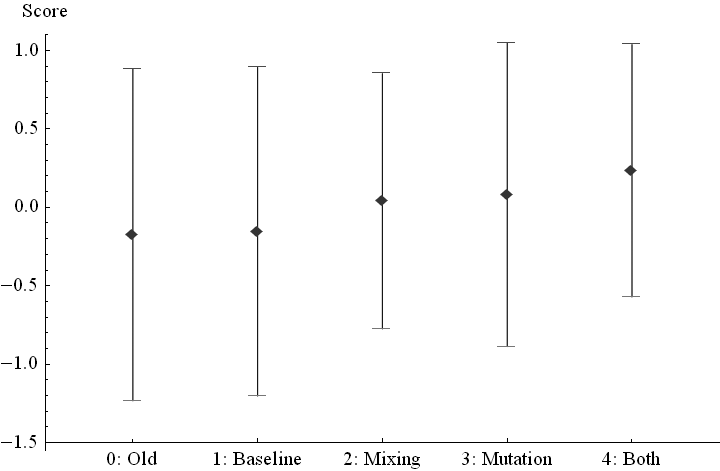}
\caption{Comparison of normalized rating score distributions between
  swarms produced under five experimental conditions (from
  \cite{sayama09ieee}). Average rating scores are shown by diamonds,
  with error bars around them showing standard deviations.}
\label{means}
\end{figure}

\begin{figure}[tbp]
\centering
\begin{tabular}{l}
(a)\\
\includegraphics[width=0.2\columnwidth]{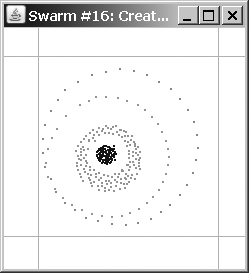}
\includegraphics[width=0.2\columnwidth]{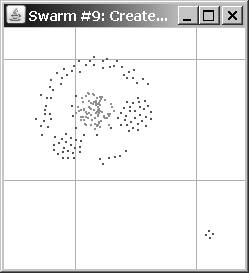}
\includegraphics[width=0.2\columnwidth]{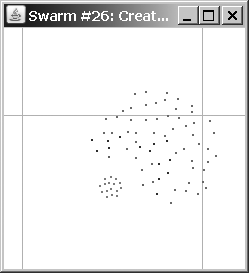}\\
~\\
(b)\\
\includegraphics[width=0.2\columnwidth]{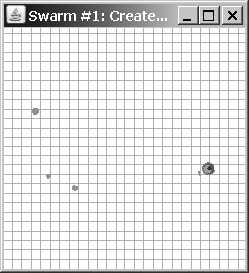}
\includegraphics[width=0.2\columnwidth]{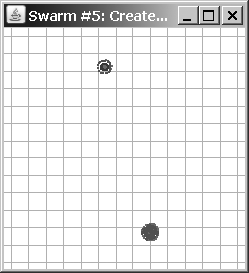}
\includegraphics[width=0.2\columnwidth]{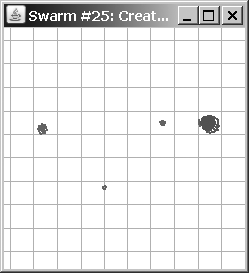}
\end{tabular}
\caption{Samples of the final swarm designs created by subjects (from
  \cite{sayama09ieee}). (a) Best three that received the highest
  rating scores. (b) Worst three that received the lowest rating
  scores.}
\label{best-worst}
\end{figure}

The result is shown in Fig.~\ref{means}. There was a difference in the
average rating scores between designs created using NIEC and HIEC
(conditions 0 and 4), and the rating scores were higher when more
evolutionary operators were made available. Several final designs
produced through the experiment are shown in Fig. \ref{best-worst}
(three with the highest scores and three with the lowest scores),
which indicate that highly evaluated swarms tended to maintain
coherent, clear structures and motions without dispersal, while those
that received lower ratings tended to disperse so that their behaviors
are not appealing to students.

To detect statistical differences between experimental conditions, a
one-way ANOVA was conducted. The result of the ANOVA is summarized in
Table \ref{anova}. Statistically significant variation was found
between the conditions ($p < 0.005$). Tukey's and Bonferroni's
post-hoc tests detected a significant difference between conditions 0
(NIEC) and 4 (HIEC), which supports our hypothesis that the HIEC is
more effective at producing final designs of higher quality than
NIEC. The post-hoc tests also detected a significant difference
between conditions 1 (HIEC without mixing or mutation operators) and 4
(HIEC). These results indicate that the more active role a designer
plays in the interactive design process, and the more diverse
evolutionary operators she has at her disposal, the more effectively
she can guide the evolutionary design of self-organizing swarms.

\begin{table}[tbp]
\label{anova}
\caption{Results of one-way ANOVA on the rating scores for five
  conditions obtained in experiment 2 (from
  \cite{bush11}). Significant difference is shown with an asterisk.}
\begin{tabular}{lccccc}
\hline
Source of variation & Degrees of freedom & Sum of squares & Mean square & $F$ & $F$-test $p$-value \\
\hline
Between groups & 4 & 14.799 & 3.700 & 4.11 & 0.003*\\
\hline
Within groups & 583 & 525.201 & 0.901 & & \\
\hline
Total & 587 & 540 & & & \\
\hline
\end{tabular}
\end{table}

\section{Automated Approach}

The second approach we took was motivated by the following question:
{\em Do we really need human users in order to guide designs of
  self-organizing swarms?} This question might sound almost
paradoxical, because designing an artifact implies the existence of a
designer by definition. However, this argument is quite similar to the
``watchmaker'' argument claimed by the English theologist William
Paley (as well as by many other leading scientists in the past)
\cite{dawkins1996blind}. Now that we know that the blind evolutionary
process did ``design'' quite complex, intricate structures and
functions of biological systems, it is reasonable to assume that it
should be possible to create automatic processes that can
spontaneously produce various creative self-organizing swarms without
any human intervention.

\begin{figure}[tbp]
\centering
\includegraphics[width=\textwidth]{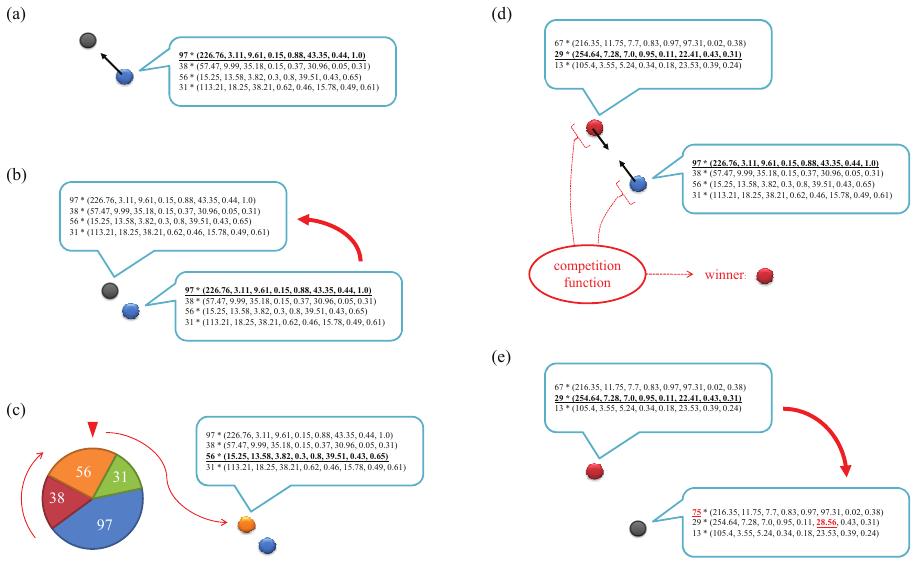}
\caption{How particle interactions work in the revised Swarm Chemistry
  (from \cite{sayama11a}). (a) There are two categories of
  particles, active (blue) and passive (gray). An active particle
  holds a recipe of the swarm in it (shown in the call-out). Each row
  in the recipe represents one kinetic parameter set. The underline
  shows which kinetic parameter set the particle is currently using
  (i.e., which kinetic type it is differentiated into). (b) A recipe
  is transmitted from an active particle to a passive particle when
  they collide, making the latter active. (c) The activated particle
  differentiates randomly into a type specified by one of the kinetic
  parameter sets in the recipe given to it. (d) A recipe is
  transmitted between active particles of different types when they
  collide. The direction of recipe transmission is determined by a
  competition function that picks one of the two colliding particles
  as a source (and the other as a target) of transmission based on
  their properties. (e) The recipe can mutate when transmitted with
  small probability.}
\label{transmission}
\end{figure}

In order to make the swarms capable of spontaneous evolution within a
simulated world, we implemented several major modifications to Swarm
Chemistry \cite{sayama10,sayama11a,sayama11b}, as follows:
\begin{enumerate}
\item There are now two categories of particles, active (moving and
  interacting kinetically) and passive (remaining still and inactive).
  An active particle holds a recipe of the swarm (a list of kinetic
  parameter sets) (Fig.~\ref{transmission}(a)).
\item A recipe is transmitted from an active particle to a passive
  particle when they collide, making the latter active
  (Fig.~\ref{transmission}(b)).
\item The activated particle differentiates randomly into one of the
  multiple types specified in the recipe, with probabilities
  proportional to their ratio in it (Fig.~\ref{transmission}(c)).
\item Active particles randomly and independently re-differentiate
  with small probability, $r$, at every time step ($r=0.005$ for all
  simulations presented in this chapter).
\item A recipe is transmitted even between two active particles of
  different types when they collide. The direction of recipe
  transmission is determined by a competition function that picks one
  of the two colliding particles as a source (and the other as a
  target) of transmission based on their properties
  (Fig.~\ref{transmission}(d)).
\item The recipe can mutate when transmitted, as well as spontaneously
  at every time step, with small probabilities, $p_t$ and $p_s$,
  respectively (Fig.~\ref{transmission}(e)). In a single recipe
  mutation event, several mutation operators are applied, including
  duplication of a kinetic parameter set (5\% per set), deletion of a
  kinetic parameter set (5\% per set), addition of a random kinetic
  parameter set (10\% per event; increased to 50\% per event in later
  experiments), and a point mutation of kinetic parameter values (10\%
  per parameter).
\end{enumerate}

These extensions made the model capable of showing morphogenesis and
self-repair \cite{sayama10} and autonomous ecological/evolutionary
behaviors of self-organized ``super-organisms'' made of a number of
swarming particles \cite{sayama11a,sayama11b}. We note here that there
was a technical problem in the original implementation of collision
detection in an earlier version of evolutionary Swarm Chemistry
\cite{sayama11a}, which was fixed in the later implementation
\cite{sayama11b}.

In addition, in order to make evolution occur, we needed to confine
the particles in a finite environment in which different recipes
compete against each other. We thus conducted all the simulations with 10,000 particles contained in a
finite, $5,000 \times 5,000$ square space (in arbitrary units; for
reference, the maximal perception radius of a particle was 300). A
``pseudo''-periodic boundary condition was applied to the boundaries
of the space. Namely, particles that cross a boundary reappear from
the other side of the space just like in conventional periodic
boundary conditions, but they do not interact across boundaries with
other particles sitting near the other side of the space. In other
words, the periodic boundary condition applies only to particle
positions, but not to their interaction forces. This specific choice
of boundary treatment was initially made because of its simplicity of
implementation, but it proved to be a useful boundary condition that
introduces a moderate amount of perturbations to swarms while
maintaining their structural coherence and confining them in a finite
area.

In the simulations, two different initial conditions were used: a {\em
  random} initial condition made of 9,900 inactive particles and 100
active particles with randomly generated one-type recipes distributed
over the space, and a {\em designed} initial condition consisted of
9,999 inactive particles distributed over the space, with just one
active particle that holds a pre-designed recipe positioned in the
center of the space. Specifically, recipes of ``swinger'', ``rotary''
and ``walker-follower'' (shown in Fig. \ref{patterns}) patterns were
used.

\subsection{Exploring experimental conditions}

Using the evolutionary Swarm Chemistry model described above, we
studied what kind of experimental conditions (competition functions
and mutation rates) would be most successful in creating
self-organizing complex patterns \cite{sayama11a}.

The first experiment was to observe the basic evolutionary dynamics of
the model under low mutation rates ($p_t = 10^{-3}$, $p_s =
10^{-5}$). Random and designed (``swinger'') initial conditions were
used. The following four basic competition functions were implemented
and tested:
\begin{itemize}
\item{\em faster}: The faster particle wins.
\item{\em slower}: The slower particle wins.
\item{\em behind}: The particle that hit the other one from behind
  wins.  Specifically, if a particle exists within a 90-degree angle
  opposite to the other particle's velocity, the former particle is
  considered a winner.
\item{\em majority}: The particle surrounded by more of the same type
  wins. The local neighborhood radius used to count the number of
  particles of the same type was 30. The absolute counts were used for
  comparison.
\end{itemize}

\begin{figure}[tbp]
\centering
\includegraphics[width=\textwidth]{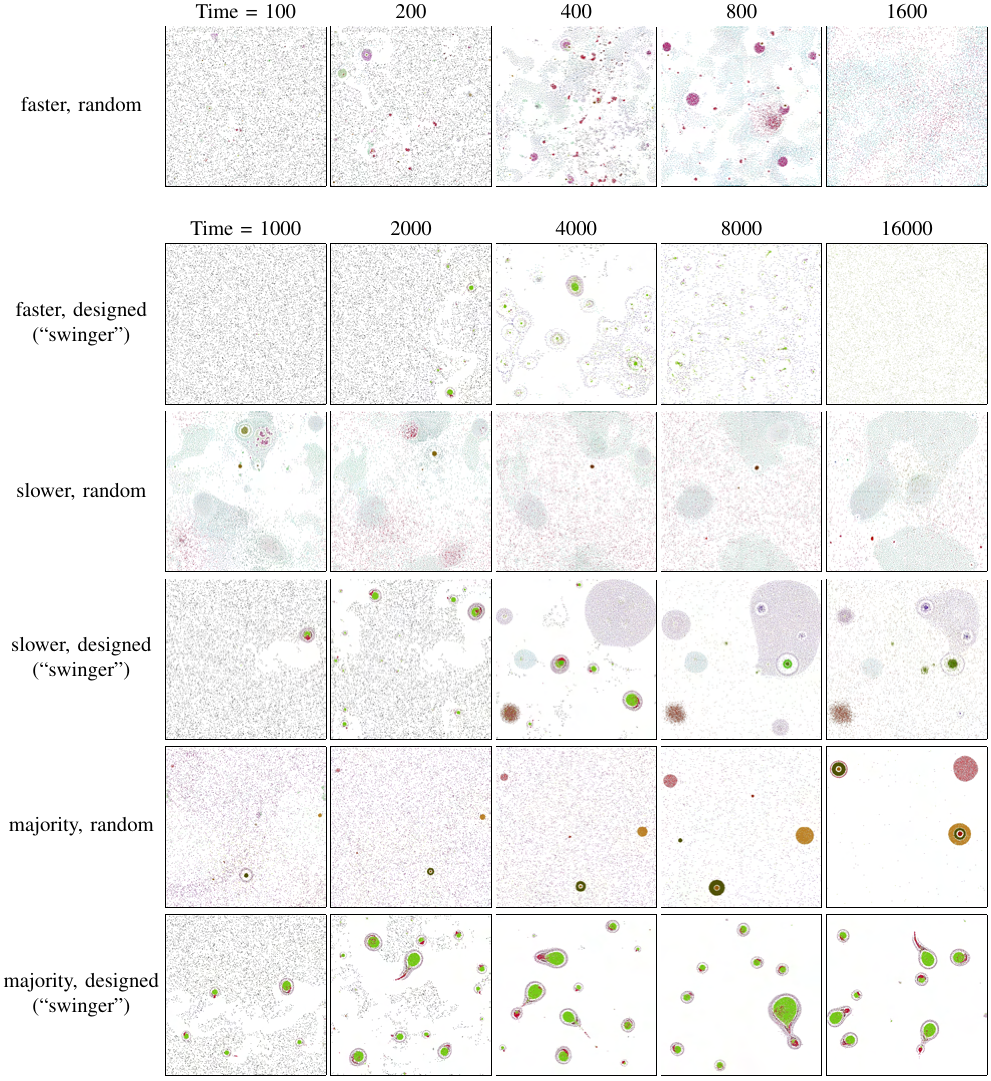}
\caption{Evolutionary processes observed in the evolutionary Swarm
  Chemistry model (from \cite{sayama11a}). Each image shows a snapshot
  of the space in a simulation, where dots with different colors
  represent particles of different types. Labels on the left indicates
  the competition function and the initial condition used in each
  case. Snapshots were taken at logarithmic time intervals.}
\label{exp1}
\end{figure}

Results are shown in Fig.~\ref{exp1}. The results with the ``behind''
competition function were very similar to those with the ``faster''
competition function, and therefore omitted from the figure. In
general, growth and replication of macroscopic structures were
observed at early stages of the simulations. The growth was
accomplished by recruitment of inactive particles through
collisions. Once a cluster of active particles outgrew maximal size
beyond which they could not maintain a single coherent structure
(typically determined by their perception range), the cluster
spontaneously split into multiple smaller clusters, naturally
resulting in the replication of those structures. These growth and
replication dynamics were particularly visible in simulations with
designed initial conditions. Once formed, the macroscopic structures
began to show ecological interactions by themselves, such as chasing,
predation and competition over finite resources (i.e., particles), and
eventually the whole system tended to settle down in a static or
dynamic state where only a small number of species were
dominant. There were some evolutionary adaptations also observed
(e.g., in faster \& designed (``swinger''); second row in
Fig.~\ref{exp1}) even with the low mutation rates used.

It was also observed that the choice of competition functions had
significant impacts on the system's evolutionary dynamics. Both the
``faster'' and ``behind'' competition functions always resulted in an
evolutionary convergence to a homogeneous cloud of fast-moving, nearly
independent particles. In contrast, the ``slower'' competition
function tended to show very slow evolution, often leading to the
emergence of crystallized patterns. The ``majority'' competition
function turned out to be most successful in creating and maintaining
dynamic behaviors of macroscopic coherent structures over a long
period of time, yet it was quite limited regarding the capability of
producing evolutionary innovations. This was because any potentially
innovative mutation appearing in a single particle would be lost in
the presence of local majority already established around it.

Based on the results of the previous experiment, the following five
more competition functions were implemented and tested. The last three
functions that took recipe length into account were implemented in the
hope that they might promote evolution of increasingly more complex
recipes and therefore more complex patterns:
\begin{itemize}
\item{\em majority (probabilistic)}: The particle surrounded by more
  of the same type wins. This is essentially the same function as the
  original ``majority'', except that the winner is determined
  probabilistically using the particle counts as relative
  probabilities of winning.
\item{\em majority (relative)}: The particle that perceives the higher
  density of the same type within its own perception range wins. The
  density was calculated by dividing the number of particles of the
  same type by the total number of particles of any kind, both counted
  within the perception range. The range may be different and
  asymmetric between the two colliding particles.
\item{\em recipe length}: The particle with a recipe that has more
  kinetic parameter sets wins.
\item{\em recipe length then majority}: The particle with a recipe
  that has more kinetic parameter sets wins. If the recipe length is
  equal between the two colliding particles, the winner is selected
  based on the ``majority'' competition function.
\item{\em recipe length $\times$ majority}: A numerical score is
  calculated for each particle by multiplying its recipe length by the
  number of particles of the same type within its local neighborhood
  (radius = 30). Then the particle with a greater score wins.
\end{itemize}

\begin{figure}[tbp]
\centering
\includegraphics[width=\textwidth]{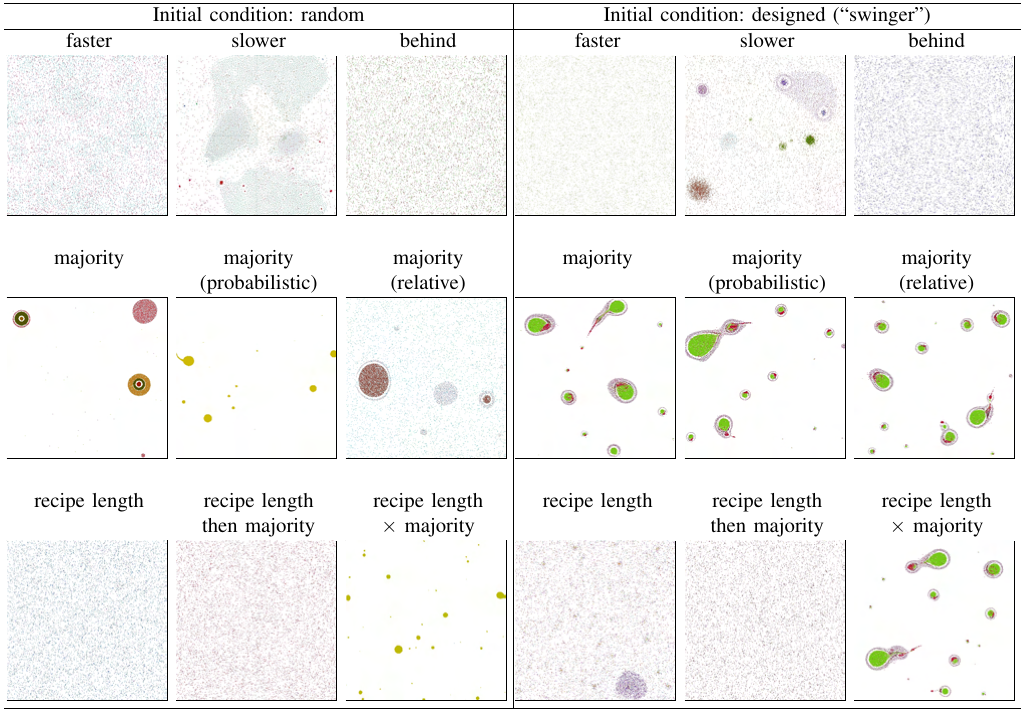}
\caption{Comparison between several different competition functions
  (from \cite{sayama11a}). The nine cases on the left hand side
  started with random initial conditions, while the other nine on the
  right hand side started with designed initial conditions with the
  ``swinger'' recipe. Snapshots were taken at time = 20,000 for all
  cases.}
\label{exp2}
\end{figure}

Results are summarized in Fig.~\ref{exp2}. As clearly seen in the
figure, the majority-based rules are generally good at maintaining
macroscopic coherent structures, regardless of minor variations in
their implementations. This indicates that interaction between
particles, or ``cooperation'' among particles of the same type to
support one another, is the key to creating and maintaining
macroscopic structures. Experimental observation of a number of
simulation runs gave an impression that the ``majority (relative)''
competition function would be the best in this regard, therefore this
function was used in all of the following experiments.

In the meantime, the ``recipe length'' and ``recipe length then
majority'' competition functions did not show any evolution toward
more complex forms, despite the fact that they would strongly promote
evolution of longer recipes. What was occurring in these conditions
was an evolutionary accumulation of ``garbage'' kinetic parameter sets
in a recipe, which did not show any interesting macroscopic
structure. This is qualitatively similar to the well-known observation
made in Tierra \cite{ray92}.

The results described above suggested the potential of evolutionary
Swarm Chemistry for producing more creative, continuous evolutionary
processes, but none of the competition functions showed notable
long-term evolutionary changes yet. We therefore increased the
mutation rates to a 100 times greater level than those in the
experiments above, and also introduced a few different types of
exogenous perturbations to create a dynamically changing environment
(for more details, see \cite{sayama11a}). This was informed by our
earlier work on evolutionary cellular automata
\cite{salzberg04a,salzberg04b}, which demonstrated that such dynamic
environments may make evolutionary dynamics of a system more
variation-driven and thus promote long-term evolutionary changes.

\begin{figure}[tbp]
\centering
\begin{tabular}{cccccc}
\fbox{\includegraphics[width=16.5mm]{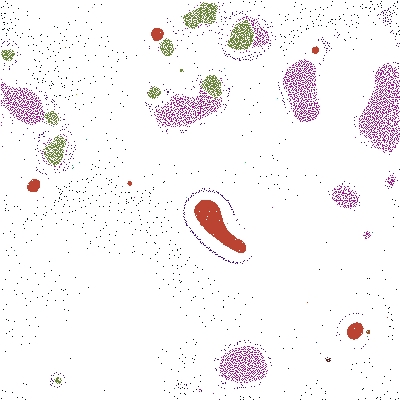}} &
\fbox{\includegraphics[width=16.5mm]{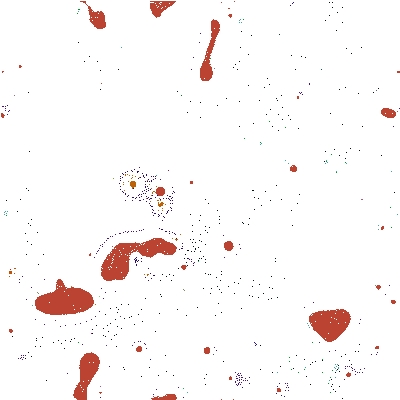}} &
\fbox{\includegraphics[width=16.5mm]{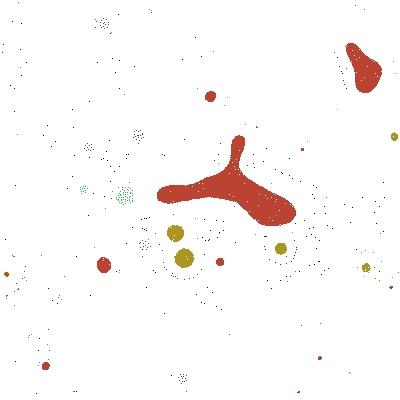}} &
\fbox{\includegraphics[width=16.5mm]{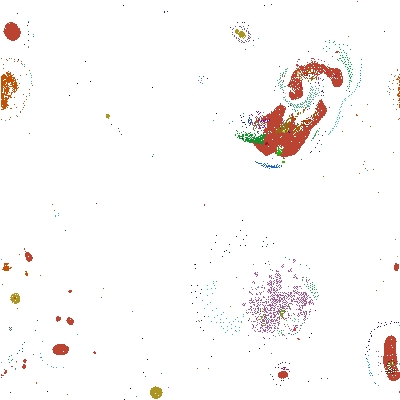}} &
\fbox{\includegraphics[width=16.5mm]{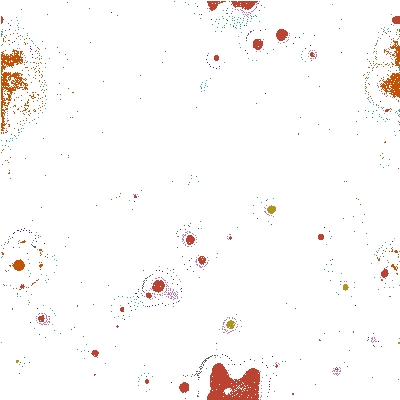}} &
\fbox{\includegraphics[width=16.5mm]{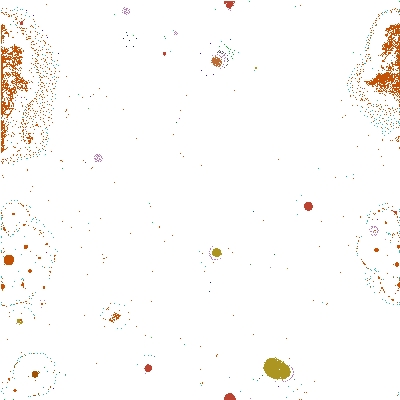}} \\
 & & & & & \\
\fbox{\includegraphics[width=16.5mm]{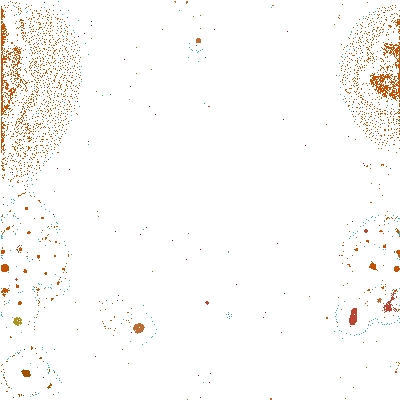}} &
\fbox{\includegraphics[width=16.5mm]{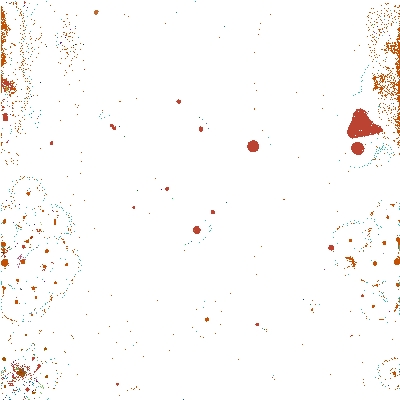}} &
\fbox{\includegraphics[width=16.5mm]{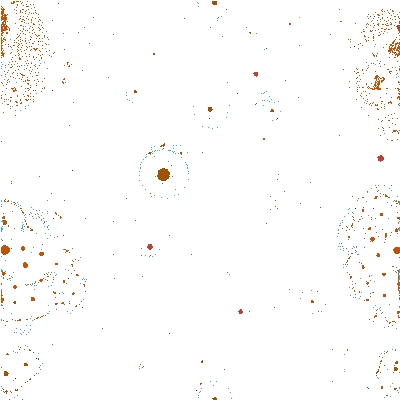}} &
\fbox{\includegraphics[width=16.5mm]{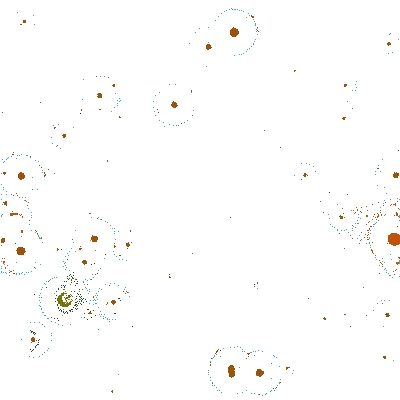}} &
\fbox{\includegraphics[width=16.5mm]{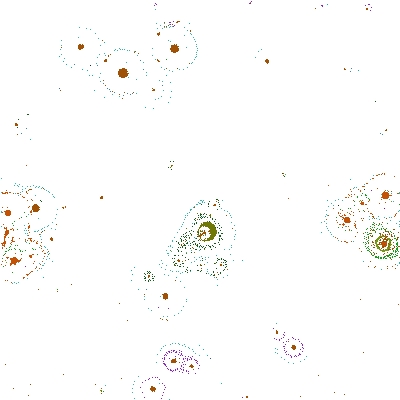}} &
\fbox{\includegraphics[width=16.5mm]{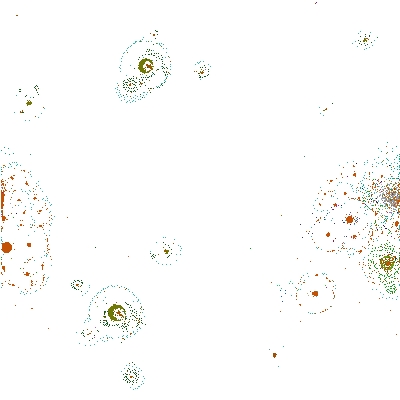}} \\
\end{tabular}
\caption{An example of long-term evolutionary behavior seen under
  dynamic environmental conditions with high mutation rates. Snapshots
  were taken at constant time intervals (2,500 steps) to show
  continuous evolutionary changes.}
\label{exp5}
\end{figure}

With these additional changes, some simulation runs finally
demonstrated continuous changes of dominant macroscopic structures
over a long period of time (Fig. \ref{exp5}). A fundamental difference
between this and earlier experiments was that the perturbation
introduced to the environment would often break the ``status quo''
established in the swarm population, making room for further
evolutionary innovations to take place. A number of unexpected,
creative swarm designs spontaneously emerged out of these simulation
runs, fulfilling our intension to create automated evolutionary design
processes. Videos of sample simulation runs can be found on our
YouTube channel (http://youtube.com/ComplexSystem).

\subsection{Quantifying observed evolutionary dynamics}

The experimental results described above were quite promising, but
they were evaluated only by visual inspection with no objective
measurements involved. To address the lack of quantitative
measurements, we developed and tested two simple measurements to
quantify the degrees of evolutionary exploration and macroscopic
structuredness of swarm populations \cite{sayama11b}, assuming that
the evolutionary process of swarms would look interesting and creative
to human eyes if it displayed patterns that are clearly visible and
continuously changing. These measurements were developed so that they
can be easily calculated a posteriori from a sequence of snapshots
(bitmap images) taken in past simulation runs, without requiring
genotypic or genealogical information that was typically assumed
available in other proposed metrics \cite{bedau92,bedau99,nehaniv00}.

\begin{sidewaysfigure}
\vspace*{5in}
\includegraphics[width=\textheight]{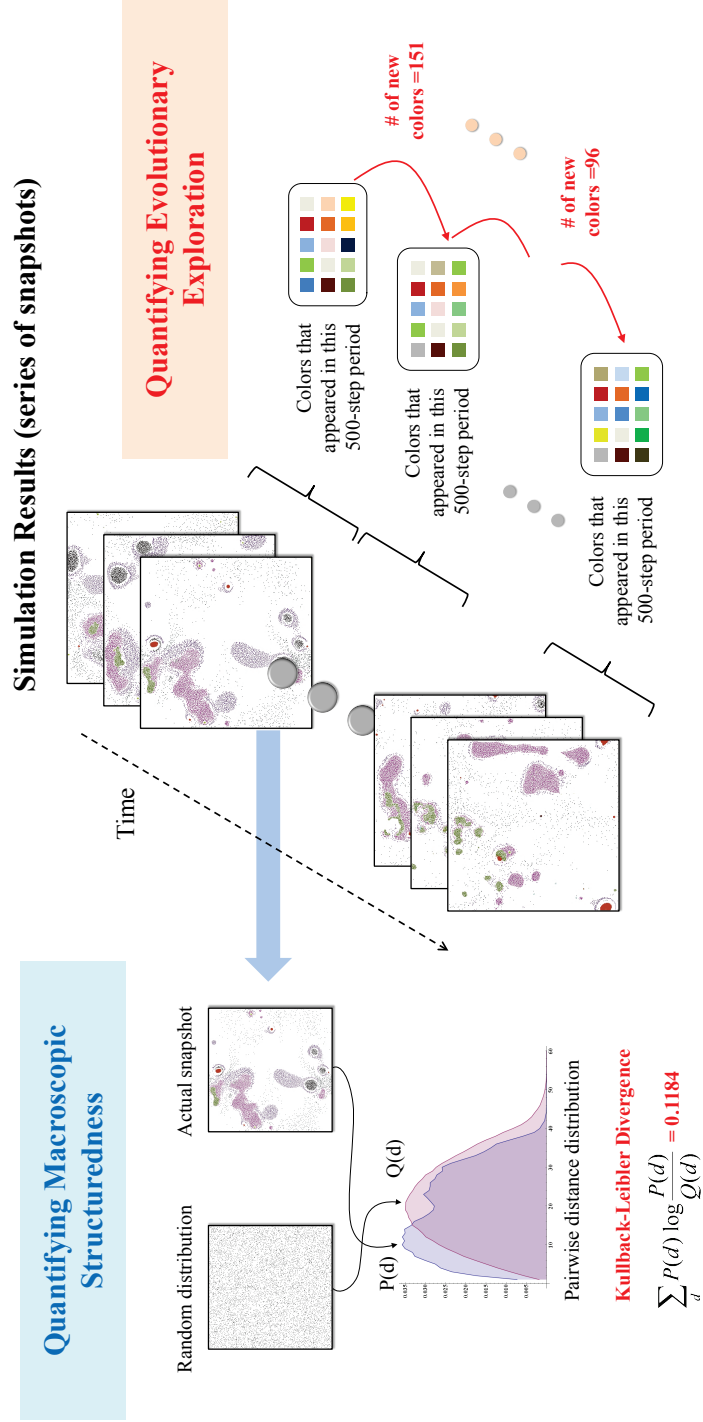}
\caption{Methods to quantify evolutionary exploration (right) and
  macroscopic structuredness (left) directly from a sequence of
  snapshots (bitmap images, center).}
\label{quantifying}
\end{sidewaysfigure}

Evolutionary exploration was quantified by counting the number of new
RGB colors that appeared in a bitmap image of the simulation snapshot
at a specific time point for the first time during each simulation run
(Fig. \ref{quantifying}, right). Since different particle types are
visualized with different colors in Swarm Chemistry, this measurement
roughly represents how many new particle types emerged during the last
time segment. Macroscopic structuredness was quantified by measuring a
Kullback-Leibler divergence \cite{kullback51} of a pairwise particle
distance distribution from that of a theoretical case where particles
are randomly and homogeneously spread over the entire space
(Fig. \ref{quantifying}, left). Specifically, each snapshot bitmap
image was first analyzed and converted into a list of coordinates
(each representing the position of a particle, or a colored pixel),
then a pair of coordinates were randomly sampled from the list 100,000
times to generate an approximate pairwise particle distance
distribution in the bitmap image. The Kullback-Leibler divergence of
the approximate distance distribution from the homogeneous case is
larger when the swarm is distributed in a less homogeneous manner,
forming macroscopic structures.

\begin{table}[tbp]
\centering
\caption{Four conditions used for the final experiment to quantify
  evolutionary dynamics.}
\label{conditions}
\begin{tabular}{c|ccc}
\hline
Name & Mutation rate & Environmental & Collision detection \\
 & & perturbation & algorithm \\
\hline
{\em original-low} & low & off & original \\
{\em original-high} & high & on & original \\
{\em revised-low} & low & off & revised \\
{\em revised-high} & high & on & revised \\
\hline
\end{tabular}
\end{table}

\begin{figure}[tbp]
\centering
\includegraphics[width=0.6\textwidth]{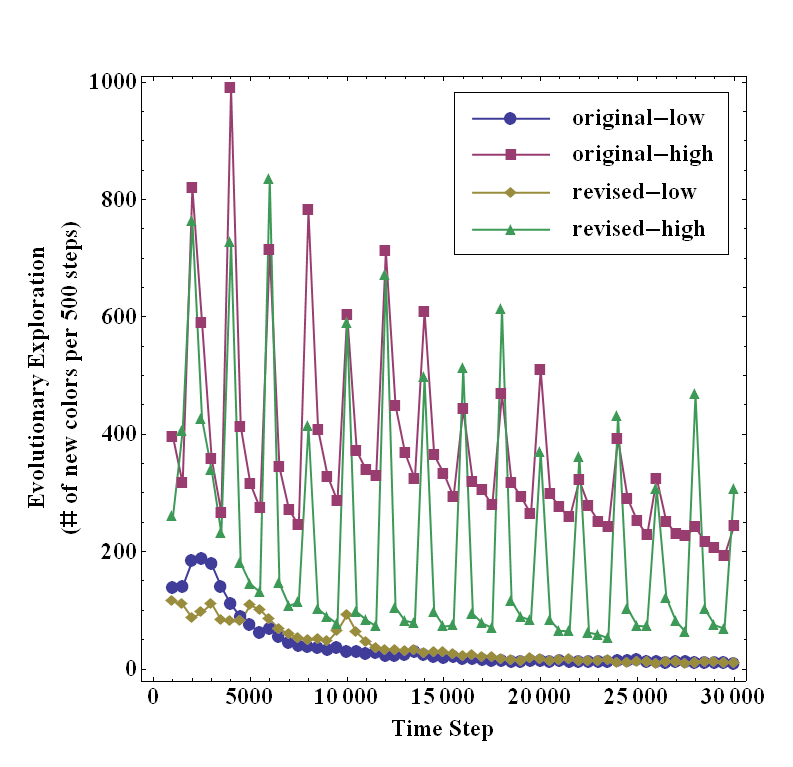}
\caption{Temporal changes of the evolutionary exploration measurement
  (i.e., number of new colors per 500 time steps) for four different
  experimental conditions, calculated from snapshots of simulation
  runs taken at 500 time step intervals (from \cite{sayama11b}). Each
  curve shows the average result over 12 simulation runs (3
  independent runs $\times$ 4 different initial conditions given in
  \cite{sayama11a}). Sharp spikes seen in ``high'' conditions were due
  to dynamic exogenous perturbations.}
\label{EE}
\end{figure}

\begin{figure}[tbp]
\centering
\includegraphics[width=0.6\textwidth]{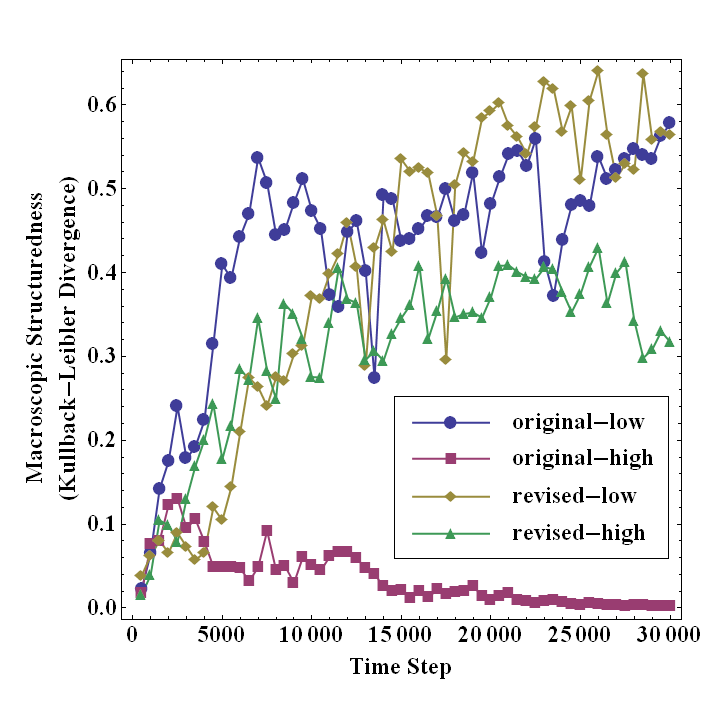}
\caption{Temporal changes of the macroscopic structuredness
  measurement (i.e., Kullback-Leibler divergence of the pairwise
  particle distance distribution from that of a purely random case)
  for four different experimental conditions, calculated from
  snapshots of simulation runs taken at 500 time step intervals (from
  \cite{sayama11b}). Each curve shows the average result over 12
  simulation runs (3 independent runs ~ 4 different initial
  conditions). The ``original-high'' condition loses macroscopic
  structures while other conditions successfully maintain them.}
\label{MS}
\end{figure}

\begin{figure}[tbp]
\centering
\includegraphics[width=0.6\textwidth]{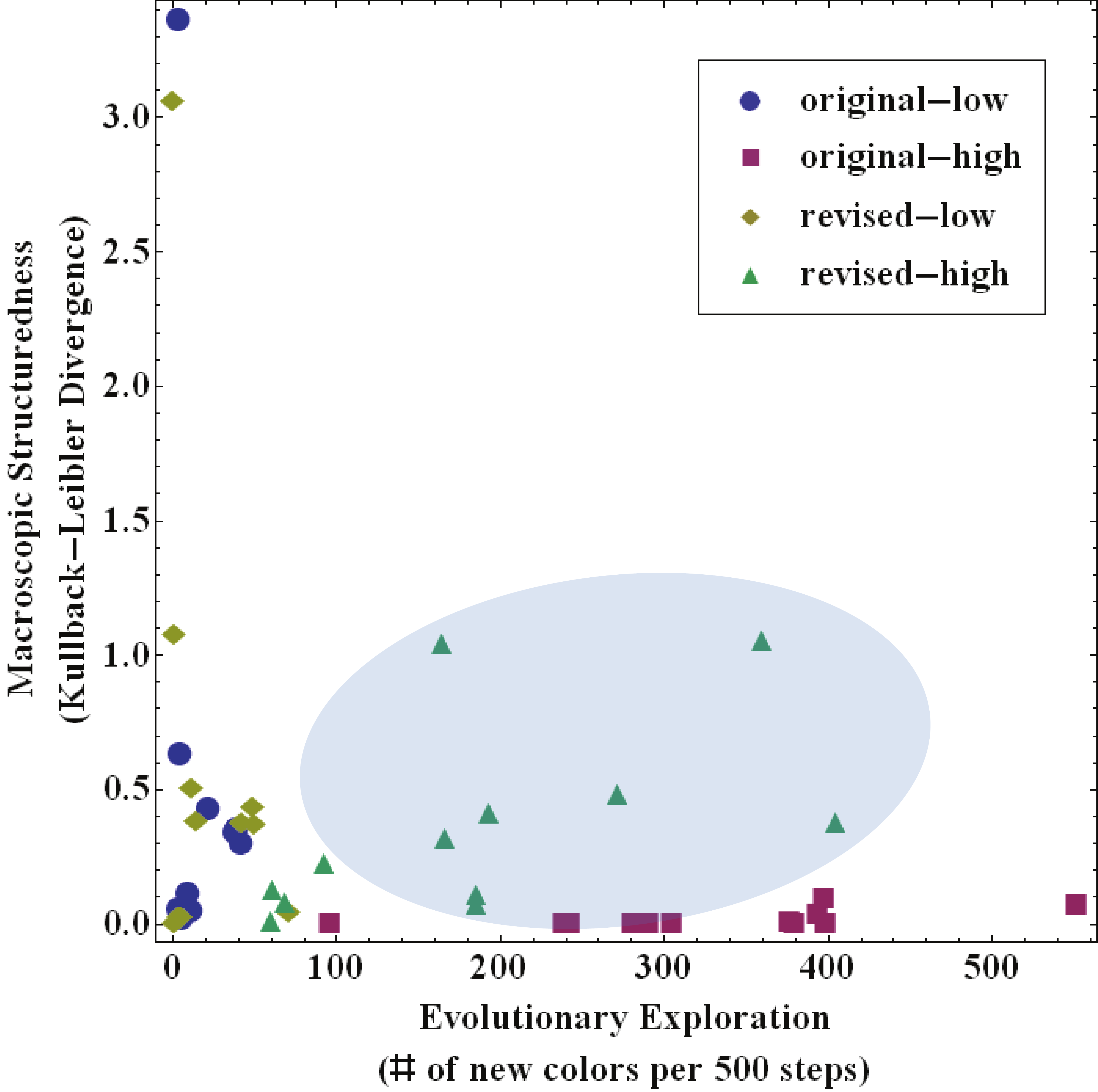}
\caption{Evolutionary exploration and macroscopic structuredness
  averaged over $t = 10,000-30,000$ for each independent simulation
  run (from \cite{sayama11b}, with slight modifications). Each marker
  represents a data point taken from a single simulation run. It is
  clearly observed that the ``revised-high'' condition (shaded in
  light blue) most successfully achieved high evolutionary exploration
  without losing macroscopic structuredness.}
\label{EEMS}
\end{figure}

We applied these measurements to simulation runs obtained under each
of the four conditions shown in Table \ref{conditions}. Results are
summarized in Figs. \ref{EE} and \ref{MS}. Figure \ref{EE} clearly
shows the high evolutionary exploration occurring under the conditions
with high mutation rates and environmental perturbations. In the
meantime, Figure \ref{MS} shows that the ``original-high'' condition
had a tendency to destroy macroscopic structures by allowing swarms to
evolve toward simpler, homogeneous forms. Such degradation of
structuredness over time was, as mentioned earlier, due to a technical
problem in the previous implementation of collision detection
\cite{sayama11a,sayama11b} that mistakenly depended on perception
ranges of particles. The ``revised'' conditions used a fixed collision
detection algorithm. This modification was found to have an effect to
maintain macroscopic structures for a prolonged period of time
(Fig. \ref{MS}). Combining these results together (Fig. \ref{EEMS}),
we were able to detect automatically that the ``revised-high''
condition was most successful in producing interesting designs,
maintaining macroscopic structures without losing evolutionary
exploration. This conclusion also matched subjective observations made
by human users.

\section{Conclusions}

In this chapter, we have reviewed our recent work on two complementary
approaches for guiding designs of self-organizing heterogeneous
swarms. The common design challenge addressed in both approaches was
the lack of explicit criteria for what constitutes a ``good'' design
to produce. In the first approach, this challenge was solved by having
a human user as an active initiator of evolutionary design
processes. In the second approach, the criteria were replaced by
low-level competition functions (similar to laws of physics) that
drive spontaneous evolution of swarms in a virtual ecosystem.

The core message arising from both approaches is the unique power of
evolutionary processes for designing self-organizing complex
systems. It is uniquely powerful because evolution does not require
any macroscopic plan, strategy or global direction for the design to
proceed. As long as the designer---this could be either an intelligent
entity or a simple unintelligent machinery---can make local decisions
at microscopic levels, the process drives itself to various novel
designs through unprescribed evolutionary pathways. Designs made
through such open-ended evolutionary processes may have a potential to
be more creative and innovative than those produced through
optimization for explicit selection criteria.

We conclude this chapter with a famous quote by Richard Feynman. At
the time of his death, Feynman wrote on a blackboard, {\em ``What I
  cannot create, I do not understand.''} This is a concise yet
profound sentence that beautifully summarizes the role and importance
of constructive understanding (i.e., model building) in scientific
endeavors, which hits home particularly well for complex systems
researchers. But research on evolutionary design of complex systems,
including ours discussed here, has illustrated that the logical
converse of the above quote is not necessarily true. That is,
evolutionary approaches make this also possible---{\em ``What I do not
  understand, I can still create.''}

\section*{Acknowledgments}

We thank the following collaborators and students for their
contributions to the research presented in this chapter: Shelley
Dionne, Craig Laramee, David Sloan Wilson, J. David Schaffer, Francis
Yammarino, Benjamin James Bush, Hadassah Head, Tom Raway, and Chun
Wong. This material is based upon work supported by the US National
Science Foundation under Grants No.\ 0737313 and 0826711, and also by
the Binghamton University Evolutionary Studies (EvoS) Small Grant (FY
2011).

\bibliographystyle{apalike}
\bibliography{sayama}

\begin{thebibliography}{}

\bibitem[Anderson, 2006]{anderson2006creation}
Anderson, C. (2006).
\newblock Creation of desirable complexity: strategies for designing
  selforganized systems.
\newblock In {\em Complex Engineered Systems}, pages 101--121. Springer.

\bibitem[Banzhaf, 2000]{banzhaf2000interactive}
Banzhaf, W. (2000).
\newblock Interactive evolution.
\newblock {\em Evolutionary Computation}, 1:228--236.

\bibitem[Bar-Yam, 2003]{bar2003dynamics}
Bar-Yam, Y. (2003).
\newblock {\em Dynamics of complex systems}.
\newblock Westview Press.

\bibitem[Bedau and Brown, 1999]{bedau99}
Bedau, M.~A. and Brown, C.~T. (1999).
\newblock Visualizing evolutionary activity of genotypes.
\newblock {\em Artificial Life}, 5(1):17--35.

\bibitem[Bedau and Packard, 1992]{bedau92}
Bedau, M.~A. and Packard, N.~H. (1992).
\newblock Measurement of evolutionary activity, teleology, and life.
\newblock In {\em Artificial Life II}, pages 431--461. Addison-Wesley.

\bibitem[Bentley, 1999]{bentley1999evolutionary}
Bentley, P. (1999).
\newblock {\em Evolutionary design by computers}.
\newblock Morgan Kaufmann.

\bibitem[Bentley and O'Reilly, 2001]{bentley}
Bentley, P.~J. and O'Reilly, U.-M. (2001).
\newblock Ten steps to make a perfect creative evolutionary design system.
\newblock In {\em GECCO 2001 Workshop on Non-Routine Design with Evolutionary
  Systems}.

\bibitem[Boccara, 2010]{boccara2010modeling}
Boccara, N. (2010).
\newblock {\em Modeling complex systems}.
\newblock Springer.

\bibitem[Bush and Sayama, 2011]{bush11}
Bush, B.~J. and Sayama, H. (2011).
\newblock Hyperinteractive evolutionary computation.
\newblock {\em Evolutionary Computation, IEEE Transactions on}, 15(3):424--433.

\bibitem[Camazine, 2003]{camazine2003self}
Camazine, S. (2003).
\newblock {\em Self-organization in biological systems}.
\newblock Princeton University Press.

\bibitem[Cheng et~al., 2005]{cheng05}
Cheng, J., Cheng, W., and Nagpal, R. (2005).
\newblock Robust and self-repairing formation control for swarms of mobile
  agents.
\newblock In {\em AAAI}, volume~5, pages 59--64.

\bibitem[Couzin et~al., 2002]{couzin02}
Couzin, I.~D., Krause, J., James, R., Ruxton, G.~D., and Franks, N.~R. (2002).
\newblock Collective memory and spatial sorting in animal groups.
\newblock {\em Journal of theoretical biology}, 218(1):1--11.

\bibitem[Dawkins, 1996]{dawkins1996blind}
Dawkins, R. (1996).
\newblock {\em The blind watchmaker: Why the evidence of evolution reveals a
  universe without design}.
\newblock WW Norton \& Company.

\bibitem[Dittrich et~al., 2001]{dittrich01}
Dittrich, P., Ziegler, J., and Banzhaf, W. (2001).
\newblock Artificial chemistries\a review.
\newblock {\em Artificial life}, 7(3):225--275.

\bibitem[Doursat et~al., 2012]{MEbookChap1}
Doursat, R., Sayama, H., and Michel, O. (2012).
\newblock Morphogenetic engineering: Reconciling self-organization and
  architecture.
\newblock In {\em Morphogenetic Engineering}, pages 1--24. Springer.

\bibitem[Hemelrijk and Kunz, 2005]{hemelrijk05}
Hemelrijk, C.~K. and Kunz, H. (2005).
\newblock Density distribution and size sorting in fish schools: an
  individual-based model.
\newblock {\em Behavioral Ecology}, 16(1):178--187.

\bibitem[Kullback and Leibler, 1951]{kullback51}
Kullback, S. and Leibler, R.~A. (1951).
\newblock On information and sufficiency.
\newblock {\em The Annals of Mathematical Statistics}, 22(1):79--86.

\bibitem[Kunz and Hemelrijk, 2003]{kunz03}
Kunz, H. and Hemelrijk, C.~K. (2003).
\newblock Artificial fish schools: collective effects of school size, body
  size, and body form.
\newblock {\em Artificial life}, 9(3):237--253.

\bibitem[Minai et~al., 2006]{minai2006complex}
Minai, A.~A., Braha, D., and Bar-Yam, Y. (2006).
\newblock {\em Complex engineered systems: A new paradigm}.
\newblock Springer.

\bibitem[Nehaniv, 2000]{nehaniv00}
Nehaniv, C.~L. (2000).
\newblock Measuring evolvability as the rate of complexity increase.
\newblock In {\em Artificial Life VII Workshop Proceedings}, pages 55--57.

\bibitem[Newman and Sayama, 2008]{newman08}
Newman, J.~P. and Sayama, H. (2008).
\newblock Effect of sensory blind zones on milling behavior in a dynamic
  self-propelled particle model.
\newblock {\em Physical Review E}, 78(1):011913.

\bibitem[Ottino, 2004]{ottino2004engineering}
Ottino, J.~M. (2004).
\newblock Engineering complex systems.
\newblock {\em Nature}, 427(6973):399--399.

\bibitem[Pahl et~al., 2007]{pahl2007engineering}
Pahl, G., Wallace, K., and Blessing, L. (2007).
\newblock {\em Engineering design: a systematic approach}, volume 157.
\newblock Springer.

\bibitem[Ray, 1992]{ray92}
Ray, T.~S. (1992).
\newblock An approach to the synthesis of life.
\newblock In {\em Artificial Life II}, pages 371--408. Addison-Wesley.

\bibitem[Reynolds, 1987]{reynolds87}
Reynolds, C.~W. (1987).
\newblock Flocks, herds and schools: A distributed behavioral model.
\newblock {\em ACM SIGGRAPH Computer Graphics}, 21(4):25--34.

\bibitem[Salzberg et~al., 2004]{salzberg04a}
Salzberg, C., Antony, A., and Sayama, H. (2004).
\newblock Evolutionary dynamics of cellular automata-based self-replicators in
  hostile environments.
\newblock {\em BioSystems}, 78(1):119--134.

\bibitem[Salzberg and Sayama, 2004]{salzberg04b}
Salzberg, C. and Sayama, H. (2004).
\newblock Complex genetic evolution of artificial self-replicators in cellular
  automata.
\newblock {\em Complexity}, 10(2):33--39.

\bibitem[Sayama, 2007]{sayama07}
Sayama, H. (2007).
\newblock Decentralized control and interactive design methods for large-scale
  heterogeneous self-organizing swarms.
\newblock In {\em Advances in Artificial Life}, pages 675--684. Springer.

\bibitem[Sayama, 2009]{sayama09}
Sayama, H. (2009).
\newblock Swarm chemistry.
\newblock {\em Artificial Life}, 15(1):105--114.

\bibitem[Sayama, 2010]{sayama10}
Sayama, H. (2010).
\newblock Robust morphogenesis of robotic swarms.
\newblock {\em Computational Intelligence Magazine, IEEE}, 5(3):43--49.

\bibitem[Sayama, 2011]{sayama11a}
Sayama, H. (2011).
\newblock Seeking open-ended evolution in swarm chemistry.
\newblock In {\em Artificial Life (ALIFE), 2011 IEEE Symposium on}, pages
  186--193. IEEE.

\bibitem[Sayama, 2012]{sayama12}
Sayama, H. (2012).
\newblock Swarm-based morphogenetic artificial life.
\newblock In {\em Morphogenetic Engineering}, pages 191--208. Springer.

\bibitem[Sayama et~al., 2009]{sayama09ieee}
Sayama, H., Dionne, S., Laramee, C., and Wilson, D.~S. (2009).
\newblock Enhancing the architecture of interactive evolutionary design for
  exploring heterogeneous particle swarm dynamics: An in-class experiment.
\newblock In {\em Artificial Life, 2009. ALife'09. IEEE Symposium on}, pages
  85--91. IEEE.

\bibitem[Sayama and Wong, 2011]{sayama11b}
Sayama, H. and Wong, C. (2011).
\newblock Quantifying evolutionary dynamics of swarm chemistry.
\newblock In {\em Advances in Artificial Life, ECAL 2011: Proceedings of the
  Eleventh European Conference on Artificial Life}, pages 729--730.

\bibitem[Shneiderman et~al., 2009]{shneiderman}
Shneiderman, B., Plaisant, C., Cohen, M., and Jacobs, S. (2009).
\newblock {\em Designing the User Interface: Strategies for Effective
  Human-Computer Interaction (5th Edition)}.
\newblock Prentice Hall.

\bibitem[Sole and Goodwin, 2008]{sole2008signs}
Sole, R. and Goodwin, B. (2008).
\newblock {\em Signs of life: How complexity pervades biology}.
\newblock Basic books.

\bibitem[Takagi, 2001]{takagi2001interactive}
Takagi, H. (2001).
\newblock Interactive evolutionary computation: Fusion of the capabilities of
  ec optimization and human evaluation.
\newblock {\em Proceedings of the IEEE}, 89(9):1275--1296.

\bibitem[Unemi, 2003]{unemi}
Unemi, T. (2003).
\newblock Simulated breeding--a framework of breeding artifacts on the
  computer.
\newblock {\em Kybernetes}, 32(1/2):203--220.

\end{thebibliography}

\end{document}